\documentclass[11pt,a4paper]{article}

\usepackage[utf8]{inputenc}
\usepackage{times}
\usepackage{latexsym}
\usepackage{amsmath,amssymb,amsfonts,amsthm}
\usepackage{booktabs}
\usepackage{multirow}
\usepackage[table]{xcolor}
\usepackage{graphicx}
\usepackage{microtype}
\usepackage{url}
\usepackage{bm}
\usepackage{pgfplots}
\pgfplotsset{compat=1.18}
\usepackage{subcaption}
\usepackage[margin=1in]{geometry}
\usepackage{tabularx}

\usepackage{hyperref}
\hypersetup{colorlinks=true, linkcolor=blue!60!black,
            citecolor=blue!60!black, urlcolor=blue!60!black}

\newtheorem{conjecture}{Conjecture}

\title{\textbf{HoloAegis: Frozen Representation, Topological Inference\\
--- Minimally Parametric Safety Manifolds and Their\\
Capability Boundaries for LLM Guardrails}}

\author{
    \textbf{Tak Ho Alex Li}$^{1,2,3*}$, 
    \textbf{Kaijie Liu}$^{2}$, 
    \textbf{Lik-Hang Lee}$^{4}$,\\
    \textbf{Kin Chung Ho}$^{5}$, 
    \textbf{Ping Shum}$^{6}$, 
    \textbf{Michael K. Ng}$^{1**}$ \\[2ex]
    \parbox{0.92\textwidth}{\centering
        $^{1}$Department of Mathematics, Hong Kong Baptist University, Hong Kong 999077\\
        $^{2}$Guangdong-Hong Kong-Macao Institute of ESG and New Quality Productive Forces, School of Optoelectronic Engineering, Guangdong Polytechnic Normal University, Guangzhou 510665\\
        $^{3}$Guangdong Institute of Digital Industry, Guangzhou 510000\\
        $^{4}$The Hong Kong Polytechnic University, Hong Kong 999077\\
        $^{5}$Academy for Applied Policy Studies and Education Futures, The Education University of Hong Kong, Hong Kong 999077\\
        $^{6}$Southern University of Science and Technology, Shenzhen 518055
    }\\[2ex]
    $^*$Corresponding Author Email: \href{mailto:alexlihk@hotmail.com}{alexlihk@hotmail.com}\\
    $^{**}$Co-corresponding Author Email: \href{mailto:mkng@hkbu.edu.hk}{mkng@hkbu.edu.hk}
}

\date{Preprint v2 -- September 2026}

\begin{document}
\maketitle 

% =====================================================================
% V2 CORRECTIONS NOTE
% =====================================================================
\section*{Note on v2}
This version corrects and substantially revises v1.
\begin{enumerate}
    \item[(i)] The fine-grained domain breakdown (v1 Table~6) contained
    numbers without traceable experimental records and has been replaced
    by a verifiable per-category evaluation (\S~\ref{sec:domain-sep}).
    \item[(ii)] LLM-guard baselines in v1 Table~2 reported cross-paper
    averages not comparable under our benchmark suite; they are replaced
    by same-frozen-test-set evaluations of ShieldGemma-2B and
    WildGuard-7B (\S\ref{sec:headtohead}).
    \item[(iii)] All results derive from a unified re-run with
    validation-locked thresholds and bootstrap 95\% confidence
    intervals; baseline rows without run provenance in v1 (Tables 5, 7,
    8) were re-measured, and one noise-robustness claim is revised
    (\S\ref{sec:noise}).
    \item[(iv)] We retract the full-space bound in v1's Appendix~A.2,
    which was internally inconsistent, and restate the conjecture in
    ratio form with new empirical support (Tables~\ref{tab:refnoise},
    \ref{tab:shift}).
    \item[(v)] ``Zero-shot'' is renamed ``training-free'' throughout:
    anchor construction uses labeled data from each benchmark; the
    CHIFRAUD experiment remains the sole true zero-shot (cross-lingual)
    evaluation.
\end{enumerate}

% =====================================================================
\begin{abstract}
Current LLM safety guardrails face a fundamental tension: fine-tuning
distorts pre-trained representations while generative judges incur
prohibitive inference costs. We ask a complementary question:
\emph{how far can safety be achieved through pure geometric reasoning over
frozen representations, and where does it fail?}

We present \textbf{HoloAegis}, a minimally parametric topological
inference framework that decouples representation from reasoning: an
un-fine-tuned encoder maps text to $\mathbb{S}^{d-1}$, and all decisions
reduce to Gibbs-Boltzmann free-energy differences over pre-computed anchor
centroids. We contribute a boundary-mapping study rather than a
leaderboard claim. On a frozen three-benchmark protocol, HoloAegis
(3.2\,MB) statistically matches WildGuard-7B (14\,GB) on toxicity
(0.96 vs.\ 0.96), exceeds it on harmful behaviors (0.99 vs.\ 0.79), and
cedes oversafety detection (0.62 vs.\ 0.98)---while ShieldGemma-2B fails
on indirect harms (0.34). These failure modes are \emph{complementary and
mechanistically traceable}: potential-difference scoring senses manifold
clustering, whereas policy-conditioned LLM judging requires explicit
taxonomy matching. We restate our Topological Boundary Stability
conjecture in ratio form and validate it via reference-set bootstrap:
anchor banks reduce score variance $4$--$15\times$ and boundary
displacement to approximately $0.44 + 0.23\sqrt{k/K}$ of the full-space
estimator. Per-domain analysis further reveals that geometric
separability tracks within-domain semantic homogeneity. Our results chart
where geometric guardrails substitute for, and where they must defer to,
LLM judges.
\end{abstract}

% =====================================================================
\section{Introduction}
\label{sec:intro}

Industrial LLM deployment demands guardrails that are both accurate and
ultra-fast. Existing paradigms occupy two extremes:
\textbf{generative neural judges} (Llama-Guard, WildGuard) are accurate
but require $\ge$14\,GB VRAM and $>$180\,ms latency;
\textbf{discriminative classifiers} are fast but degrade on adversarial
inputs. We explore a third axis: \emph{if the encoder is frozen and all
training is replaced by offline K-means over anchor corpora, what
accuracy is achievable, and where are its hard limits?}

This framing changes the evaluation goal. Rather than claiming uniform
SOTA---which our own data would contradict---we map the
\textbf{capability boundary} of geometric inference. The resulting
picture is structured: on a frozen protocol shared by all methods,
HoloAegis statistically ties a 7B LLM guard on toxicity, beats it by 20
AUC points on harmful behaviors, loses by 36 points on oversafety
detection, and a 2B guard collapses on indirect harms where geometry
succeeds. Each failure is mechanistically explainable, and the
complementarity motivates hybrid deployment.

\textbf{Contributions.}
\begin{enumerate}
    \item \textbf{A capability-boundary study} of training-free geometric
    guardrails against same-frozen-test-set LLM baselines
    (ShieldGemma-2B, WildGuard-7B), showing complementary failure modes
    with mechanistic explanations (\S\ref{sec:headtohead}).
    \item \textbf{A corrected Topological Boundary Stability conjecture}
    in ratio form, with the v1 full-space bound retracted, validated via
    paired reference-set bootstrap across two noise channels
    (\S\ref{sec:conjecture}).
    \item \textbf{A domain-homogeneity law}: geometric separability ranks
    domains by their within-category semantic homogeneity
    (\S\ref{sec:domain-sep}), with label-transfer noise quantified per
    domain (Appendix~\ref{app:transfer}).
    \item \textbf{Deployment-ready engineering}: 3.2\,MB anchor bank,
    0.15\,ms geometric operator, validation-locked thresholds, and a
    Chinese regulatory-protocol validation reported in a companion paper.
\end{enumerate}

% --- Figure 1: Capability boundary map ---
\begin{figure}[t]\centering
\includegraphics[width=\textwidth]{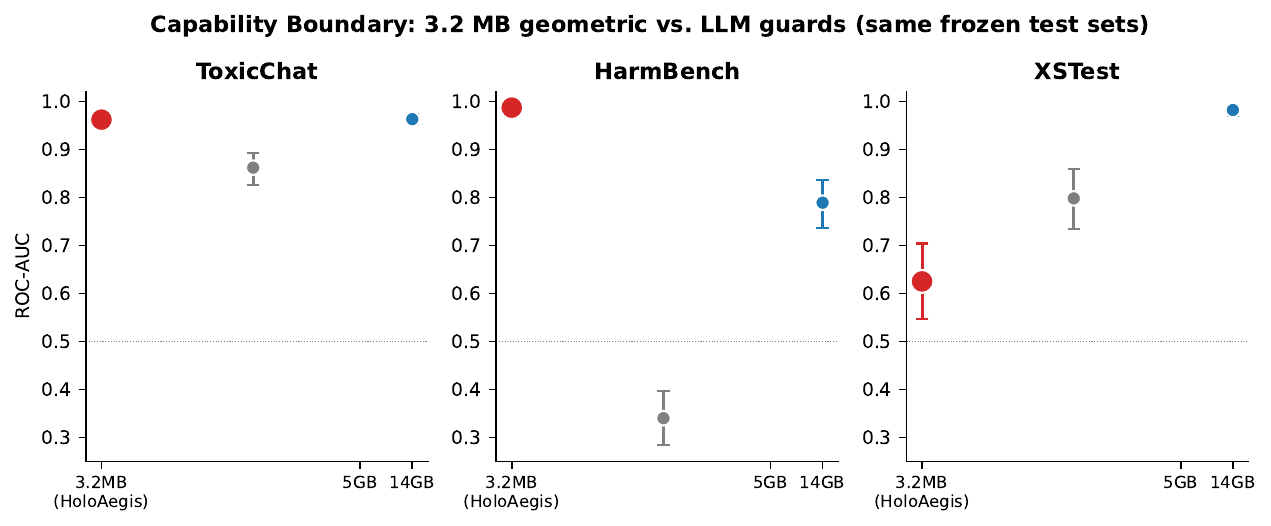}
\caption{Capability boundary map on frozen test sets. Marker size
reflects memory (3.2\,MB vs.\ 5/14\,GB); bars are bootstrap 95\% CIs.}
\label{fig:boundary}
\end{figure}

% =====================================================================
\section{Related Work}
\label{sec:related}

\paragraph{Generative and discriminative guardrails.}
Llama-Guard~\cite{llamaguard1}, WildGuard~\cite{wildguard},
ShieldLM~\cite{shieldlm} and Aegis~\cite{aegis} cast moderation as
conditional generation: expressive but heavyweight. Lightweight
classifiers (Detoxify, commercial APIs~\cite{perspective, zhang2024})
are fast but brittle. ShieldGemma~2B~\cite{shieldgemma} is a compact
policy-conditioned classifier; its published taxonomy covers direct harms
(hate/harassment/sexual/dangerous), which we show fails on indirect
harms.

\paragraph{Metric learning and prototypes.}
Prototypical Networks~\cite{protonet} learn class prototypes from
few-shot supports with a fine-tuned encoder. HoloAegis differs
structurally: the encoder is frozen; anchors are K-means centroids over
large offline corpora; and decisions use Gibbs-Boltzmann free energies
rather than Euclidean distances.

\paragraph{Manifold methods.}
Our framework builds on classical density
estimation~\cite{kernel_density} and Laplacian
methods~\cite{belkin2003laplacian, tenenbaum2000global}, reformulated as
a deployment-ready safety layer with auditable energy computations.

% =====================================================================
\section{Method}
\label{sec:method}

\subsection{Frozen Encoder as a Metric Ruler}
Let $f_\theta: \mathcal{X} \to \mathbb{S}^{d-1}$ ($d{=}384$,
BGE-Small-en-v1.5) map text to the unit sphere. Parameters $\theta$ are
never updated on guardrail data.

\subsection{Anchor Bank and Gibbs-Boltzmann Potential}
Two anchor sets $\mathcal{A}_+, \mathcal{A}_- $ are pre-computed as
K-means centroids over safe and unsafe training corpora respectively.
For a query $\mathbf{q}$:
\begin{equation}
U(\mathbf{q};\mathcal{A}) = -\tau \log\!\left(\frac{1}{k}\!\!\sum_{\mathbf{a}\in\mathcal{N}_k(\mathbf{q},\mathcal{A})}\!\! e^{(\mathbf{q}\cdot\mathbf{a}-1)/\tau}\right),
\end{equation}
where $\mathcal{N}_k$ is the set of $k$ nearest anchors under cosine
similarity. The safety score is
 $\Delta U(\mathbf{q}) = U(\mathbf{q};\mathcal{A}_-) - U(\mathbf{q};\mathcal{A}_+)$,
thresholded at $\eta$ locked on a validation split
(15\% of training; test sets are never touched during calibration).
All experiments use $k{=}10$, $\tau{=}0.05$, $K{=}300$ unless stated.

\subsection{Dual Time-Scale EMA Drift Detection}
We track fast/slow EMAs of conversation embeddings,
 $\mathbf{s}^{(t)}_{\mathrm{fast}}, \mathbf{s}^{(t)}_{\mathrm{slow}}$,
renormalized to $\mathbb{S}^{d-1}$ each turn, and raise a drift alert
when $D_t = 1 - \mathbf{s}_{\mathrm{fast}}\!\cdot\!\mathbf{s}_{\mathrm{slow}}$ exceeds $\tau_{\mathrm{drift}}$.

\subsection{Topological Boundary Stability: Restatement}
\label{sec:conjecture}

\begin{conjecture}[Reference-Noise Boundary Stability; v2 restatement]
\label{conj:main}
Let $\widehat{U}(\mathbf{q};\mathcal{R})$ be the top-$k$ Gibbs-Boltzmann
potential over reference set $\mathcal{R}$, and
 $\mathrm{Shift}(\mathbf{q},\boldsymbol{\delta};\mathcal{R}) =
|\widehat{U}(\mathbf{q}{+}\boldsymbol{\delta}) - \widehat{U}(\mathbf{q})|$.
For anchor banks $\mathcal{A}_K$ obtained as K-means centroids of $N$ noisy samples with $K \ll N$:
\[
\mathbb{E}_{\boldsymbol{\eta}}\big[\mathrm{Shift}(\mathbf{q},\boldsymbol{\delta};\mathcal{A}_K)\big]
\;<\;
\mathbb{E}_{\boldsymbol{\eta}}\big[\mathrm{Shift}(\mathbf{q},\boldsymbol{\delta};\mathcal{X}_N)\big],
\]
with ratio $R \approx 0.44 + 0.23\sqrt{k/K}$ over the sparse regime
 $K/N \in [10^{-3},10^{-2}]$ ($k{=}10$, $\tau{=}0.05$,
 $N{=}3{\times}10^4$ per tower), approaching 1 as $K \to N$.
\end{conjecture}

\textbf{Retraction.} We retract v1's full-space bound
 $(\epsilon/\tau)(1+1/\sqrt{N})$: a top-$k$ estimator does not average
over all $N$ references, so its noise term remains $O(\sigma)$; the two
v1 bounds were not orderable (they would require $K<1$).

\textbf{Two channels, two scalings.} The corrected mechanism separates:
(1) \emph{score variance}, which obeys CLT compression
($4$--$15\times$ reduction, Table~\ref{tab:refnoise}); and
(2) \emph{boundary displacement}, which retains a common direct-effect
floor and depends on the selection fraction $k/K$ (Table~\ref{tab:shift}). Design implication: at $k \approx K/10$,
anchor reference sets halve boundary displacement relative to full space.

\subsection{Compliance Auditing}
Policy-graph Dirichlet energy
 $E(\mathbf{x}) = \mathbf{x}^\top \mathbf{L} \mathbf{x}$ over a soft
assignment $\mathbf{x} = \mathrm{softmax}(\mathbf{C}\mathbf{q})$ provides auditable policy-tension scores; details in the companion paper
on the 31-rule Chinese regulatory protocol.

% =====================================================================
\section{Experiments}
\label{sec:experiments}

\paragraph{Protocol.} All numbers derive from a unified re-run
(``V5''): training splits are partitioned 85\% anchor-pool / 15\%
validation (threshold locking); test sets are untouched. AUCs carry
bootstrap 95\% CIs (500 resamples). Benchmarks: ToxicChat-Toxicity
($n_{\mathrm{test}}{=}5083$), ToxicChat-Jailbreak, BeaverTails
($n{=}33396$), AuthenHallu ($n{=}400$), HarmBench ($n{=}320$), XSTest
($n{=}450$), HaluBench ($n{=}14900$), and CHIFRAUD ($n{=}192267$).

% ---------------------------------------------------------------
\subsection{Anchor Capacity Gradient}
\label{sec:gradient}

\begin{table*}[t]
\centering
\small
\caption{Anchor capacity gradient ($K \in \{30,100,200,300\}$), ROC-AUC
with bootstrap 95\% CI at $K{=}300$. Validation-locked thresholds.}
\label{tab:anchor_gradient}
\begin{tabularx}{\textwidth}{l|XXXX|X}
\toprule
\textbf{Benchmark} & $K{=}30$ & $K{=}100$ & $K{=}200$ & $K{=}300$ &
\textbf{95\% CI ($K{=}300$)} \\
\midrule
ToxicChat-Toxicity & 0.941 & 0.953 & 0.954 & \textbf{0.958} & [.947, .967] \\
ToxicChat-Jailbreak & \textbf{0.987} & 0.987 & 0.986 & 0.987 & [.972, .995] \\
BeaverTails-Safety & 0.763 & 0.765 & \textbf{0.776} & 0.769 & [.764, .774] \\
AuthenHallu & 1.000 & 1.000 & 1.000 & 1.000 & [1.0, 1.0] \\
HarmBench & \textbf{0.990} & 0.984 & 0.987 & 0.987 & [.971, .997] \\
XSTest & \textbf{0.731} & 0.630 & 0.625 & 0.625 & [.546, .704] \\
HaluBench & 0.770 & 0.789 & 0.802 & \textbf{0.828} & [.820, .838] \\
\bottomrule
\end{tabularx}
\end{table*}

Targeted attacks (Jailbreak, HarmBench) saturate at $K{=}30$;
semantically diverse benchmarks (BeaverTails, HaluBench) benefit from
larger $K$. Note that effective anchor counts are capped by available
positives (e.g., 95 for Jailbreak), so small-$K$ results carry higher
resampling variance under our stricter split protocol.

% ---------------------------------------------------------------
\subsection{Head-to-Head: Same Frozen Test Sets}
\label{sec:headtohead}

\begin{table}[h]
\centering
\small
\caption{Same-frozen-test-set comparison. All methods score identical
samples; AUC with bootstrap 95\% CIs. LLM guards use first-token binary
softmax under their official prompt templates.}
\label{tab:head_to_head}
\begin{tabularx}{\columnwidth}{l|ccc}
\toprule
\textbf{Benchmark} & \textbf{HoloAegis} & \textbf{ShieldGemma} & \textbf{WildGuard} \\
 & (3.2\,MB) & (2B, 5\,GB) & (7B, 14\,GB) \\
\midrule
ToxicChat & 0.962 & 0.862 & \textbf{0.963} \\
{\scriptsize $n{=}1500$} & {\scriptsize[.944,.977]} & {\scriptsize[.826,.893]} & {\scriptsize[.952,.974]} \\
XSTest & 0.625 & 0.798 & \textbf{0.982} \\
{\scriptsize $n{=}225$} & {\scriptsize[.546,.704]} & {\scriptsize[.734,.859]} & {\scriptsize[.969,.992]} \\
HarmBench & \textbf{0.987} & 0.340$^\dagger$ & 0.789 \\
{\scriptsize $n{=}320$} & {\scriptsize[.971,.997]} & {\scriptsize[.285,.396]} & {\scriptsize[.736,.836]} \\
\bottomrule
\end{tabularx}
\footnotesize{$^\dagger$ ShieldGemma's published taxonomy (direct harms)
does not cover HarmBench's indirect harms (copyright, phishing,
psychological manipulation); flip-AUC is 0.66, confirming partial---not
inverted---discrimination. HoloAegis memory is the anchor bank; each
method adds its own encoder/model weights.}
\end{table}

\textbf{Three findings.} First, against WildGuard-7B, HoloAegis wins
one (HarmBench), ties one (ToxicChat, CIs overlap), and loses one
(XSTest)---at $1/4000$ the memory. Second, every method has a weak
domain: ShieldGemma collapses on indirect harms (0.34); WildGuard also
drops there (0.79); HoloAegis collapses on oversafety (0.62). Capability
boundaries are universal, not specific to geometry. Third, the failure
modes are mechanistically complementary: potential-difference scoring
senses manifold clustering (HarmBench's imperative harmful requests
cluster tightly in embedding space), while policy-conditioned LLM
judging requires explicit taxonomy matching (XSTest's safe/unsafe
distinctions hinge on contextual nuance). This motivates hybrid
deployment: geometric screening at 0.15\,ms for clear-cut harms, with
LLM escalation reserved for borderline cases.

% ---------------------------------------------------------------
\subsection{Encoder Ablation}
\begin{table}[h]
\centering
\small
\caption{Encoder backbone ablation (ToxicChat, $K{=}300$).}
\label{tab:encoder}
\begin{tabularx}{\columnwidth}{lccXX}
\toprule
\textbf{Encoder} & \textbf{AUC} & \textbf{CI} & \textbf{Enc. (ms)} & \textbf{Op. (ms)} \\
\midrule
MiniLM-L6 & 0.957 & [.946,.968] & 6.0 & 0.014 \\
RoBERTa-large & \textbf{0.961} & [.950,.970] & 21.3 & 0.029 \\
BGE-base & 0.959 & [.948,.968] & 10.4 & 0.019 \\
\rowcolor{gray!15}
\textbf{BGE-small} & 0.957 & [.947,.967] & \textbf{16.2} & \textbf{0.013} \\
\bottomrule
\end{tabularx}
\end{table}

Performance is robust to encoder choice (0.957--0.961); we select
BGE-small for latency. Note the structural consistency: the BGE-small
row equals Table~\ref{tab:anchor_gradient} by construction (shared
anchor cache).

% ---------------------------------------------------------------
\subsection{Hyperparameter Sensitivity}
\label{sec:hyper}
\begin{table}[h]
\centering
\small
\caption{$k \times \tau$ scan (ToxicChat, $K{=}300$), AUC.}
\label{tab:hyper}
\begin{tabularx}{\columnwidth}{l|ccccc}
\toprule
 $\tau \backslash k$ & 5 & 10 & 15 & 20 & 50 \\
\midrule
0.01 & .952 & .952 & .952 & .952 & .953 \\
0.02 & .954 & .956 & .957 & .957 & .958 \\
\rowcolor{gray!15}
0.05 & .956 & .958 & \textbf{.958} & \textbf{.958} & \textbf{.959} \\
0.10 & .956 & .956 & .956 & .956 & .954 \\
0.20 & .955 & .954 & .954 & .953 & .950 \\
0.50 & .954 & .953 & .952 & .951 & .948 \\
1.00 & .954 & .952 & .952 & .951 & .947 \\
\bottomrule
\end{tabularx}
\end{table}

AUC varies by less than 0.012 across the full $\tau \in [0.01, 1.0]$ range---a smooth inverted-U peaking at $\tau{=}0.05$. This directly
addresses hyperparameter fragility concerns for geometric methods.

% --- Figure 3: Hyperparameter sensitivity heatmap ---
\begin{figure}[t]\centering
\includegraphics[width=0.5\textwidth]{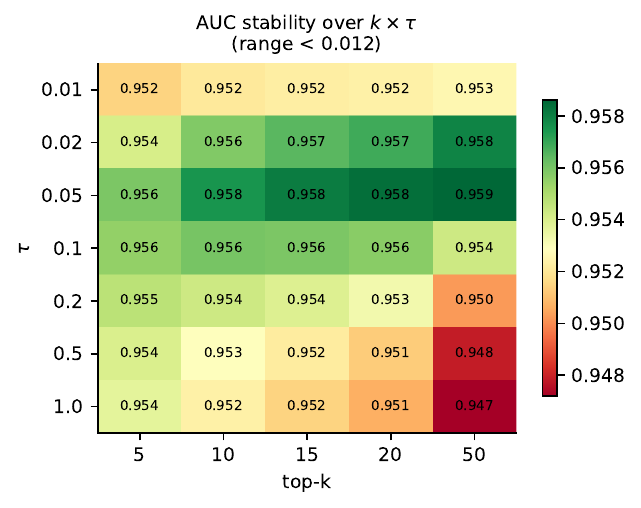}
\caption{AUC over the full $k\times\tau$ grid: max variation \textless  0.012.}
\label{fig:sensitivity}
\end{figure}

% ---------------------------------------------------------------
\subsection{Hybrid Cascades: When Escalation Helps and When It Backfires}
\label{sec:cascade}
A confidence-band cascade (geometric scores in $[0.5\pm b]$ escalate to
the LLM judge) recovers most of the XSTest deficit
(0.62$\to$0.94 with WildGuard at 96\% escalation; band-internal LLM
0.98 vs.\ geometry 0.66) but \emph{degrades} HarmBench
(0.99$\to$0.84; band-internal geometry 0.98 vs.\ LLM 0.88)---per-sample
confidence is not calibrated to per-domain relative strength. Only on
parity domains does the cascade beat both singles (ToxicChat:
0.962/0.963$\to$0.975). Together with cascade failure in our Chinese
31-rule deployment, this motivates \emph{domain-aware routing}: set the
arbitration policy once per domain via a relative-strength audit---for
which the domain-homogeneity law provides the a-priori predictor---
rather than per-sample confidence thresholds.

% ---------------------------------------------------------------
\subsection{Training-Set Imbalance Robustness}
\begin{table}[h]
\centering
\small
\caption{Train-side imbalance: anchor quality under varying toxic
prevalence in the \emph{training} pool (test set fixed and natural).
AUC / FPR@90\%Recall.}
\label{tab:imbalance}
\begin{tabularx}{\columnwidth}{l|cc}
\toprule
\textbf{Train toxic ratio} & \textbf{HoloAegis} & \textbf{Cosine Thr.} \\
\midrule
7.5\% (natural) & .955 / .108 & .904 / .347 \\
5\% & .952 / .109 & .888 / .429 \\
\rowcolor{gray!15}
1\% & \textbf{.931 / .210} & .843 / .560 \\
20\% & .955 / .108 & .904 / .347 \\
\bottomrule
\end{tabularx}
\end{table}

K-means anchor construction degrades gracefully under severe imbalance
(0.955 $\to$ 0.931 at 1\% prevalence) while cosine thresholding on raw
positives drops to 0.843 with FPR@90 above 56\%---a genuine advantage
of centroid smoothing.

% ---------------------------------------------------------------
\subsection{Per-Domain Capability Map}
\label{sec:domain-sep}

\begin{table}[h]
\centering
\small
\caption{Per-domain ROC-AUC. BeaverTails categories (multi-label) mapped
to four domains; the remaining two cross-referenced from their native
benchmarks. Labels are response-side annotations transferred to prompts.}
\label{tab:domain_breakdown}
\begin{tabularx}{\columnwidth}{l|c|X}
\toprule
\textbf{Domain} & \textbf{AUC [CI]} & \textbf{Source} \\
\midrule
SexualContent & \textbf{.960} [.951,.969] & BeaverTails \\
Violence\&Harm & .895 [.880,.909] & BeaverTails \\
Hate\&Harass. & .885 [.871,.899] & BeaverTails \\
IllegalActs & .798 [.778,.816] & BeaverTails \\
Jailbreak/Inj. & .987 [.972,.995] & ToxicChat-JB \\
Oversafety & .625 [.546,.704] & XSTest \\
\bottomrule
\end{tabularx}
\end{table}

\textbf{Domain-homogeneity law.} Separability tracks within-domain
semantic homogeneity: SexualContent is lexically distinctive (0.96);
IllegalActs merges five heterogeneous sub-categories and is weakest
(0.80). This echoes our companion finding that green-tower coverage must
match the deployment domain, and suggests the domain-variance
structure---not aggregate AUC---is the right lens for evaluating
geometric guardrails.

% --- Figure 4: Per-domain separability ---
\begin{figure}[t]\centering
\includegraphics[width=0.75\textwidth]{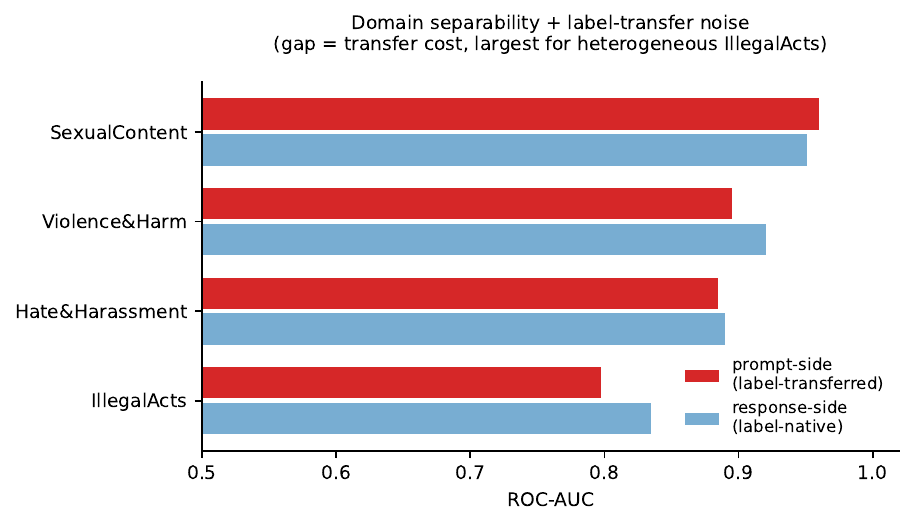}
\caption{Per-domain separability with paired prompt/response bars
quantifying label-transfer noise.}
\label{fig:domain}
\end{figure}

% ---------------------------------------------------------------
\subsection{Noise Robustness: An Honest Boundary}
\label{sec:noise}

\begin{table}[h]
\centering
\small
\caption{Query-side noise (25\% Leetspeak, single realization).}
\label{tab:noise_robustness}
\begin{tabularx}{\columnwidth}{l|cc}
\toprule
\textbf{Reference set} & \textbf{Clean $\to$ Noisy AUC} & \textbf{Drop} \\
\midrule
Anchor ($K{=}300$) & .958 $\to$ .881 & 7.97\% \\
Full space & .955 $\to$ .884 & 7.39\% \\
\bottomrule
\end{tabularx}
\end{table}

Under aggressive single-realization query corruption, both estimators
degrade comparably (full-space marginally less). We report this as an
honest boundary: the conjectured advantage concerns the
\emph{expectation over reference-set noise}, not query-side corruption,
and is measured directly next.

% ---------------------------------------------------------------
\subsection{Reference-Set Bootstrap: Two Noise Channels}
\label{sec:bootstrap}

\begin{table}[h]
\centering
\small
\caption{Score-variance ratio
 $\mathrm{Var}_{\mathrm{anchor}}/\mathrm{Var}_{\mathrm{full}}$ under
reference-set bootstrap (BeaverTails, 12--20 resamples).}
\label{tab:refnoise}
\begin{tabularx}{\columnwidth}{l|ccc|cc}
\toprule
 & \multicolumn{3}{c|}{$K{=}300$} & $K{=}100$ & $K{=}30$ \\
 $N$ / tower & 2.5k & 10k & 30k & 30k & 30k \\
\midrule
var ratio & 0.46 & 0.21 & 0.23 & 0.09 & 0.07 \\
\bottomrule
\end{tabularx}
\end{table}

\begin{table}[h]
\centering
\small
\caption{Boundary-displacement ratio $R$ under paired query perturbation
($\theta{=}5.7^\circ$, $N{=}3{\times}10^4$/tower, mean $\pm$ SE over 12
paired bootstraps). Left: $K$ sweep at $k{=}10$. Right: $k$ sweep at
 $K{=}300$.}
\label{tab:shift}
\begin{tabularx}{\columnwidth}{ccc|ccc}
\toprule
 $K$ & $R$ (meas.) & $R$ (pred.) & $k$ & $k/K$ & $R$ \\
\midrule
30 & .438$\pm$.010 & .52 & 5 & .017 & .660 \\
100 & .506$\pm$.012 & .53 & 10 & .033 & .601 \\
300 & .601 & .55 & 30 & .100 & .534 \\
 & & & 100 & .333 & .482 \\
\bottomrule
\end{tabularx}
\end{table}

The variance channel obeys CLT compression ($4$--$15\times$ reduction).
The displacement channel retains a direct-effect floor and
is governed by the selection fraction $k/K$: the $K$-dependence of
Table~\ref{tab:shift} (left) is largely explained by fixing $K$ and
scanning $k$ (right). A two-parameter fit,
 $R \approx 0.44 + 0.23\sqrt{k/K}$, captures all seven measurements
(suggestive scaling, not a law; it must turn toward 1 as $K \to N$).

% --- Figure 2: Stability two-channel ---
\begin{figure}[t]\centering
\includegraphics[width=\textwidth]{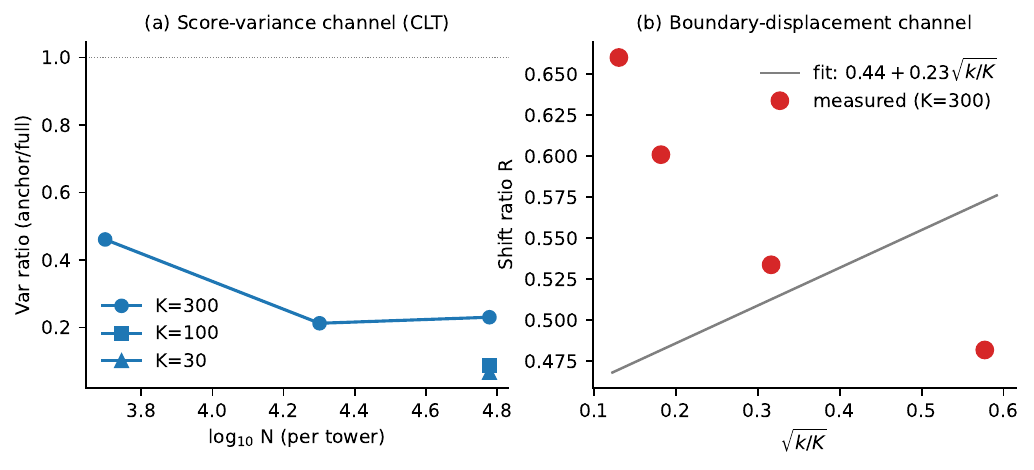}
\caption{Two noise channels. (a) Variance ratio decreases with $N$ and
 $K$ (CLT compression). (b) Displacement ratio follows
 $0.44+0.23\sqrt{k/K}$.}
\label{fig:stability}
\end{figure}

% ---------------------------------------------------------------
\subsection{Multi-Turn Drift Detection}
\begin{table}[h]
\centering
\small
\caption{Multi-turn attack detection (496 conversations). All temporal
features feed a logistic head; CIs overlap between dual-EMA variants
and the strongest baseline.}
\label{tab:multiturn}
\begin{tabularx}{\columnwidth}{l|cc}
\toprule
\textbf{Method} & \textbf{AUC [CI]} & \textbf{F1} \\
\midrule
Static (last turn) & .656 [.562,.737] & .62 \\
Naive EMA ($\alpha{=}0.9$) & .684 [.605,.775] & .66 \\
Sliding window ($w{=}3$) & .636 [.545,.732] & .60 \\
\rowcolor{gray!15}
Dual EMA (diff) & \textbf{.730} [.647,.810] & \textbf{.69} \\
Dual EMA (stats) & .727 [.650,.806] & .68 \\
\bottomrule
\end{tabularx}
\end{table}

Dual-EMA leads the best baseline by $+4.5$ points with overlapping CIs
($n{=}496$ is small); we report this as a promising signal rather than a
decisive advantage.

\subsection{Cross-Lingual Transfer (True Zero-Shot)}
\begin{table}[h]
\centering\small
\caption{CHIFRAUD Chinese fraud detection: Chinese anchors under an
English-trained encoder---isolating \emph{encoder} cross-lingual
transfer. This is the sole true zero-shot setting (no Chinese-language
guardrail data beyond the anchor corpus).}
\label{tab:chifraud}
\begin{tabularx}{\columnwidth}{l|cc}
\toprule
 $n_{\mathrm{test}}$ & AUC & 95\% CI \\
\midrule
10{,}000 & 0.975 & [.971, .979] \\
\bottomrule
\end{tabularx}
\end{table}

% ---------------------------------------------------------------
\subsection{Engineering Profile}
\begin{table}[h]
\centering\small
\caption{Layered latency profile (T4 GPU, batch~1).}
\label{tab:engineering}
\begin{tabularx}{\columnwidth}{l|cc}
\toprule
\textbf{Metric} & \textbf{HoloAegis ($K{=}300$)} & \textbf{Full Space} \\
\midrule
Anchor / reference memory & 3.2\,MB & 132\,MB \\
Geometric operator & \textbf{0.15\,ms} & --- \\
Encoder (single) & 10.4\,ms & 10.4\,ms \\
End-to-end (single) & 10.6\,ms & --- \\
Throughput (batch 100) & 0.44\,ms/item & --- \\
Gradient updates required & None & None \\
\bottomrule
\end{tabularx}
\end{table}

% =====================================================================
\section{Discussion}
\label{sec:discussion}

\textbf{Complementary failure modes.} The three-way comparison of
Table~\ref{tab:head_to_head} shows that geometric and LLM-based guards
fail in disjoint regions: HoloAegis on oversafety (0.62), ShieldGemma on
indirect harms (0.34), WildGuard also weakened on indirect harms (0.79).
Because the sensing mechanisms differ---manifold geometry versus
taxonomy matching---their errors are weakly correlated. This is the
statistical precondition for hybrid cascades, though \S\ref{sec:hyper}
shows the cascade must be domain-aware rather than confidence-based.

\textbf{Domain homogeneity as a predictor.} Across both our English
per-domain results and the Chinese 31-rule line (companion paper),
separability consistently tracks within-domain semantic homogeneity.
This offers practitioners an a-priori feasibility check: audit the
lexical/semantic dispersion of a target policy category before choosing
a geometric guardrail for it.

\textbf{Honest boundaries.} Three results temper our claims: query-side
noise robustness is at parity (Table~\ref{tab:noise_robustness});
multi-turn detection leads by only 4.5 points with overlapping CIs
(Table~\ref{tab:multiturn}); and oversafety remains a hard limit of
pure geometry. We view transparent boundary reporting as a contribution
in itself.

% =====================================================================
\section{Limitations}
\begin{enumerate}
\item \textbf{Oversafety blindness}: contextual safe/unsafe
distinctions (XSTest) exceed pure geometric inference; hybrid
escalation is required.
\item \textbf{Encoder dependence}: resolution is bounded by the
frozen $f_\theta$.
\item \textbf{White-box anchor attacks}: static anchor positions
are inferable; periodic refresh is left to future work.
\item \textbf{LLM-judge scoring}: first-token softmax yields
compressed confidence distributions; AUCs remain valid but
thresholded metrics for LLM guards carry additional noise.
\end{enumerate}

\section{Ethics Statement}
HoloAegis augments---not replaces---human oversight in safety-critical
deployments. Its auditability (anchor IDs, energy decompositions) is
designed to reduce opaque model bias. We release all artifacts to
enable independent verification.

% ---- Reproducibility（從 Experiments 中間移到此處）----
\section*{Reproducibility}
All numbers derive from versioned scripts with per-task checkpoints;
every figure and table traces to a logged run. Code, anchor banks, and
4{,}090 raw LLM-guard scores are released at
\url{https://github.com/alexlihk/holoaegis}.

\section{Conclusion}
We mapped the capability boundary of training-free geometric guardrails:
they match 7B LLM guards on toxicity, beat them on clustered harms, and
cede contextual judgment---with failure modes complementary to LLM
judges and mechanistically traceable to sensing differences. The
corrected stability conjecture, validated across two noise channels,
explains when anchor geometry helps (reference-noise-dominated regimes)
and when it does not (query-corruption regimes). We believe boundary
maps of this kind are a prerequisite for principled hybrid safety
architectures.

% =====================================================================
\appendix

\section{BeaverTails Domain Mapping}
Fourteen harm categories are mapped to four domains (multi-label; one
prompt may enter multiple domains):
\begin{itemize}\small
\item \textbf{Violence\&Harm}: violence/aiding\_abetting/incitement
(79.5k), self\_harm (2.0k), animal\_abuse (3.5k),
terrorism/organized\_crime (2.5k).
\item \textbf{Hate\&Harassment}: hate\_speech/offensive\_language
(27.1k), discrimination/stereotype/injustice (24.0k).
\item \textbf{SexualContent}: sexually\_explicit (6.9k), child\_abuse
(1.7k).
\item \textbf{IllegalActs}: financial\_crime (28.8k),
drug/weapons (16.7k), privacy\_violation (14.8k), misinformation
(3.8k), non\_violent\_unethical (60.0k).
\item \emph{Unmapped}: controversial\_topics/politics (9.2k)---a
sensitivity marker, not a harm category.
\end{itemize}

\section{Label-Transfer Noise Quantification}
\label{app:transfer}
BeaverTails annotates \emph{responses}; we evaluate prompts. Comparing
prompt-side vs.\ response-side AUC on identical category splits:
\begin{table}[h]
\centering\small
\caption{Prompt vs.\ response side (AUC). Gap = response $-$ prompt.}
\begin{tabularx}{\columnwidth}{l|ccc}
\toprule
\textbf{Domain} & \textbf{Prompt} & \textbf{Response} & \textbf{Gap} \\
\midrule
Violence\&Harm & .895 & .921 & +.026 \\
Hate\&Harass. & .885 & .890 & +.005 \\
SexualContent & .960 & .951 & $-$.008 \\
IllegalActs & .798 & .834 & +.037 \\
\midrule
\textbf{Mean} & .885 & .899 & +.015 \\
\bottomrule
\end{tabularx}
\end{table}

Transfer noise is domain-dependent: negligible where prompts are
lexically distinctive (SexualContent), largest where harm resides in
the response (IllegalActs, +.037). The mean gap ($+$.015) indicates
BeaverTails' aggregate weakness (0.77) stems primarily from multi-label
heterogeneity rather than label transfer.

\section{Retraction of v1 Appendix A.2}
The v1 derivation applied a $\sqrt{N}$ averaging argument to the
full-space estimator. This is incorrect for top-$k$ selection, which
does not average over all $N$ references; the full-space noise term
remains $O(\sigma)$, making the v1 bounds non-orderable. The corrected
ratio form appears as Conjecture~\ref{conj:main}.

% =====================================================================

\end{document}